\documentclass[pdflatex,sn-mathphys-num]{sn-jnl}

\usepackage{graphicx}%
\usepackage{multirow}%
\usepackage{amsmath,amssymb,amsfonts}%
\usepackage{amsthm}%
\usepackage{mathrsfs}%
\usepackage[title]{appendix}%
\usepackage{xcolor}%
\usepackage{textcomp}%
\usepackage{manyfoot}%
\usepackage{booktabs}%
\usepackage{algorithm}%
\usepackage{algorithmicx}%
\usepackage{algpseudocode}%
\usepackage{listings}%
\usepackage{soul}

\theoremstyle{thmstyleone}%
\usepackage[caption=false]{subfig}

\newenvironment{abbreviations}{\begin{list}{}{}}{\end{list}}
\usepackage{makecell}
\theoremstyle{thmstyletwo}%

\theoremstyle{thmstylethree}%

\begin{document}

\title[Article Title]{Examining Deployment and Refinement of the VIOLA-AI Intracranial Hemorrhage Model Using an Interactive NeoMedSys Platform }


\author*[1]{\fnm{Qinghui} \sur{Liu}}\email{qiliu@ous-hf.no}
\author[1]{\fnm{Jon E.} \sur{Nesvold}}
\author[2]{\fnm{Hanna} \sur{Raaum}}
\author[1]{\fnm{Elakkyen} \sur{Murugesu}}
\author[1]{\fnm{Martin} \sur{Røvang}}
\author[1, 4]{\fnm{Bradley J} \sur{Maclntosh}}
\author[1,3]{\fnm{Atle} \sur{Bjørnerud}}
\author[1,2]{\fnm{Karoline} \sur{Skogen}}

\affil*[1]{\orgdiv{Computational Radiology \& Artificial Intelligence Unit, Department of Physics and Computational Radiology, Clinic for Radiology and Nuclear Medicine}, \orgname{Oslo University Hospital}, \orgaddress{\street{Sognsvannsveien 20}, \postcode{0372}, \country{Norway}}}

\affil[2]{\orgdiv{Department of Physics and Computational Radiology, Clinic for Radiology and Nuclear Medicine}, \orgname{Oslo University Hospital}, \orgaddress{\country{Norway}}}

\affil[3]{\orgdiv{Department of Physics}, \orgname{University of Oslo}, \orgaddress{\country{Norway}}}

\affil[4]{\orgdiv{Department of Medical Biophysics}, \orgname{University of Toronto}, \orgaddress{\country{Canada}}}


\abstract{\textbf{Background:} There are many challenges and opportunities in the clinical deployment of AI tools in radiology. The current study describes a radiology software platform called NeoMedSys that can enable efficient deployment and refinements of AI models. We evaluated the feasibility and effectiveness of running NeoMedSys for three months in real-world clinical settings and focused on improvement performance of an in-house developed AI model (VIOLA-AI) designed for intracranial hemorrhage (ICH) detection.

\textbf{Methods:} NeoMedSys integrates tools for deploying, testing, and optimizing AI models with a web-based medical image viewer, annotation system, and hospital-wide radiology information systems. A prospective pragmatic investigation was deployed using clinical cases of patients presenting to the largest Emergency Department in Norway (site-1) with suspected traumatic brain injury (TBI) or patients with suspected stroke (site-2). We assessed ICH classification performance as VIOLA-AI encountered new data and underwent pre-planned model retraining. Performance metrics included sensitivity, specificity, accuracy, and the area under the receiver operating characteristic curve (AUC).

\textbf{Results:} NeoMedSys facilitated iterative improvements in the AI model, significantly enhancing its diagnostic accuracy. Automated bleed detection and segmentation were reviewed in near real-time to facilitate re-training VIOLA-AI. The iterative refinement process yielded a marked improvement in classification sensitivity, rising to 90.3\% (from 79.2\%), and specificity that reached 89.3\% (from 80.7\%). The bleed detection ROC analysis for the entire sample demonstrated a high area-under-the-curve (AUC) of 0.949 (from 0.873). Model refinement stages were associated with notable gains, highlighting the value of real-time radiologist feedback.

\textbf{Discussion:}
The NeoMedSys platform provided the means to evaluate and improve upon the bleed detection VIOLA-AI tool, by using prospective data collection and near real-time radiological review of incoming cases. The results of this pragmatic deployment demonstrate there is clinical value in fine-tuning an established deep-learning image analysis tool to account for local data sources and conditions. Iterative clinical feedback improved VIOLA-AI performance and informed on NeoMedSys user-experience. Ongoing efforts are to extend NeoMedSys to support a broader range of modalities, AI models, and patient groups. This in-house approach has fostered a broader interest and enthusiasm to adopt AI solutions into our clinical workflows.}




\maketitle

\section{Introduction}\label{sec1}

The recent surge of AI research in medical imaging is developing into AI applications that are becoming readily available as both in-house developed and commercial tools, but it is not yet reflected clinically \cite{ng_bridging_2020}.  Implementation and integration appear difficult especially for in-house developed algorithms often due to the lack of proper infrastructure and understanding. Around every corner, researchers and clinicians encounter technological, ethical, regulatory, and health-economic challenges deeming such processes difficult. Another critical barrier is the lack of platforms that enable efficient translation of research-based AI models into clinical prototypes. This is creating an uneven gap between the in-house and commercial tools' potential to reach clinical practice. Promoting to reduce this, investments are made to established AI departments often within established institutions such as university hospitals to elicit the adaptation of AI, both commercial and in-house, into different patient groups. 

The most common AI applications often appear in imaging related to neurology, musculoskeletal and chest, as detection, segmentation and diagnostic tools \cite{van_leeuwen_artificial_2021}. Patients with suspected intracranial hemorrhage (ICH) including traumatic brain injury (TBI) and hemorrhagic stroke are major drivers of acute head CT examination to rapidly diagnose a potential critical ICH. With a prevalence of 8-15\% and ease of detection on head CT, numerous AI models have been developed both commercially and in-house to assist radiologists with the workload \cite{af_geijerstam_mild_2003}. Collective efforts from many researchers have led to promising examples of deep learning, machine learning, and automated approaches to detect ICH, primarily using CT images \cite{yu_robust_2022, seyam_utilization_2022, schmitt_automated_2022, hopkins_mass_2022, yu_predicting_2025}. A previously published in-house developed algorithm to detect ICH, “Voxels Intersecting along Orthogonal-Levels of Attention” (VIOLA-AI) \cite{macintosh_radiological_2023} provided a starting point for a process of iterative model re-training as prospective data were collected over the course of a pragmatic clinical deployment period of time.

The purpose of the current study is to describe a platform called NeoMedSys \footnote{\url{https://www.neomedsys.io/}}, which was developed in-house to enable receiving radiology examinations directly from PACS, facilitating interactivity and annotations by clinicians, and fine-tune retraining VIOLA as new cases arrived into the platform in near real-time. We present the methods and results of this clinical deployment where knowledge and human expertise were provided dynamically to facilitate iterative refinement of VIOLA-AI. We hypothesize that ICH detection performance metrics would increase as more curated data were made available to support successive rounds of VIOLA-AI retraining and optimization. The study design was a prospective pragmatic investigation, conducted in a real-world clinical setting over a 3-month period. Cases were ingested in near real-time to evaluate performance and enable continuous refinement through clinician feedback.

\section{Methods}\label{meth}
\subsection{Setting of the Study}

\begin{figure}[!htp]
    \subfloat[]{%
      \includegraphics[clip,width=0.6269\textwidth]{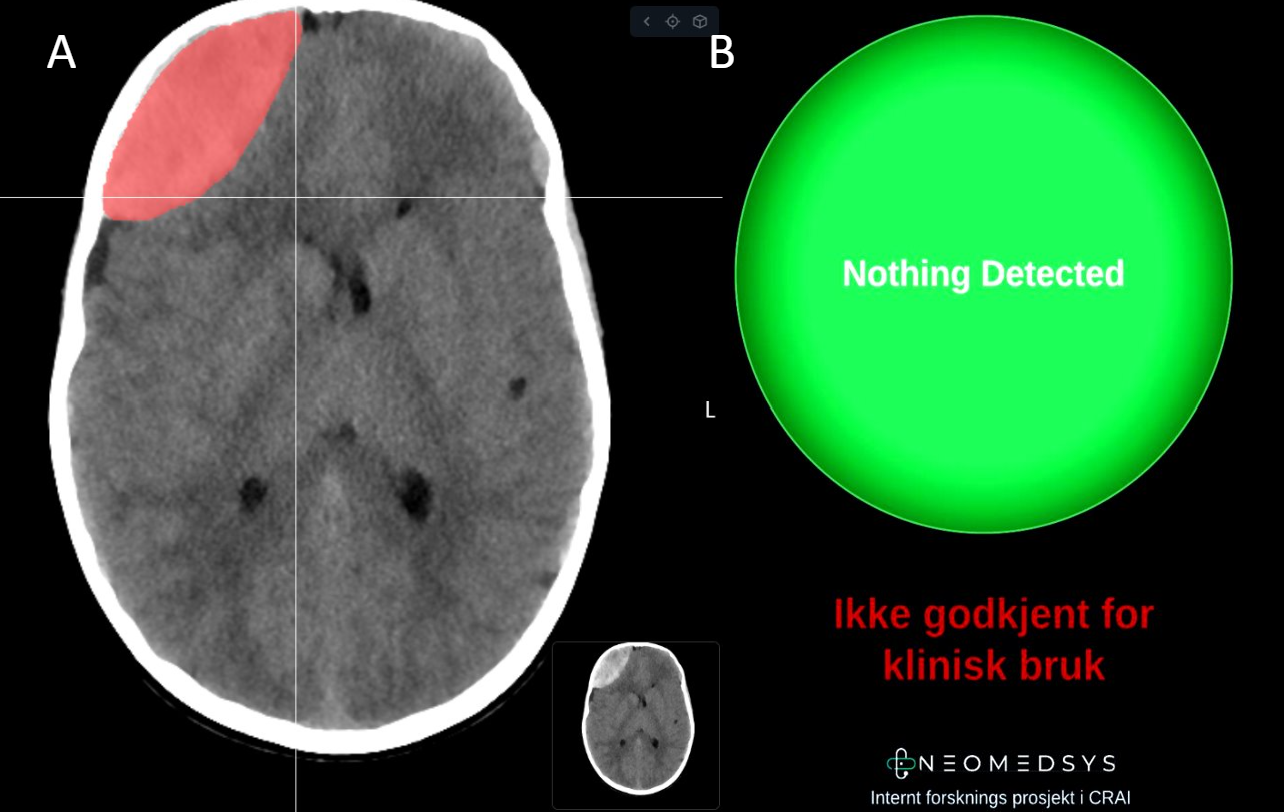}%
    }\subfloat[]{%
      \includegraphics[clip,width=0.3731\textwidth]{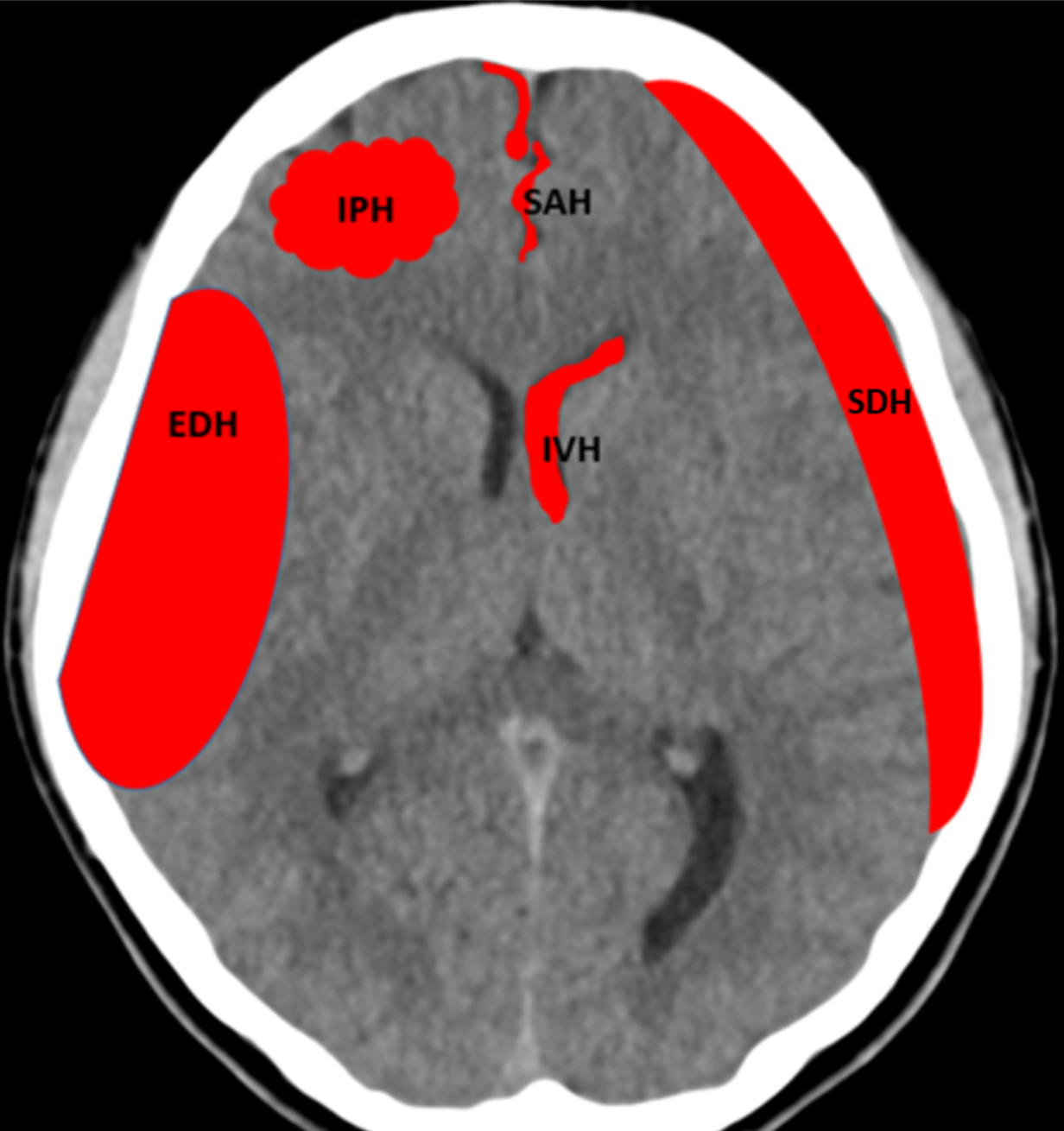}%
    }

    \subfloat[]{%
      \includegraphics[clip,width=\textwidth]{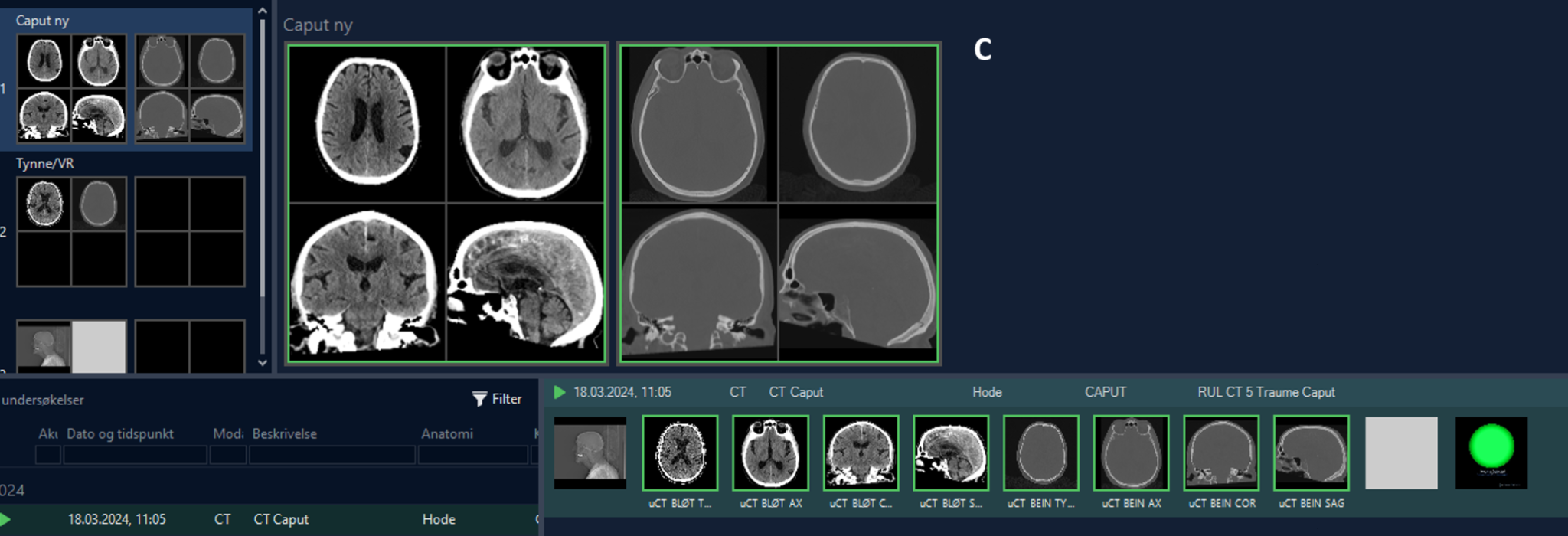}%
    }
    \caption{(a) shows A: mask presented when an ICH is detected and B: the image as a DICOM if no bleed is detected (ikke godkjent for klinisk bruk = not approved for clinical use), (b) shows the different intracranial hemorrhages, epidural hematoma (EDH), subdural hematoma (SDH), subarachnoid hematoma (SAH), intraparenchymal hematoma (IPH) and intraventricular hematoma (IVH), and (c) shows the appearance in PACS matrix window.}%
    \label{fig:intro}
\end{figure}

Over a 3 month period (from 1 June to 31 August 2024), all patients with suspected TBI and stroke patients who underwent a head CT were included in this study, and were approved by the regional ethics committee (REK 324897). Radiologists and radiographers from Oslo emergency department and Oslo University hospitals were instructed to “push” images from PACS to the NeoMedSys platform (details below). CT images came from 6 different CT scanners (Simens and GE) and were pushed to NeoMedSys as DICOM files and tagged digitally by the frontline user and processed using the VIOLA-AI model (details below) within NeoMedSys. Results of the analysis were presented as an automated VIOLA-AI report presented back in PACS consisting of a binary mask demarcating the ICH and its volume and a green circle when an ICH was not detected(Figure \ref{fig:intro}-a).
Demographic data were retrieved, including the time when: 1) CT was performed, and 2) the radiologist signed the report. Two patient groups were created based on whether an ICH was detected or not. Within the ICH detected group the clinician further classified the bleed etiology as being: epidural hematoma (EDH), acute subdural hematoma (SDH), subarachnoid hemorrhage (SAH), intraparenchymal hematoma (IPH), or intraventricular hematoma (IVH), as well as evidence of mass effect (Figure \ref{fig:intro}-b). As this was a live consecutive patient population other pathologies were also encountered and were classified as non-ICH with other pathologies i.e., tumor, non-hemorrhagic stroke, etc. All information was retrieved from PACS and the signed radiology report (Figure \ref{fig:intro}-c).

During the initial phase of the study, cases were aggregated to retrain the VIOLA-AI tool. Throughout the pragmatic clinical evaluation, new CT head cases were used for online testing purposes. 

\subsection{NeoMedSys Platform}
\begin{figure}[!htp]
    \centerline{\includegraphics[width=\textwidth ]{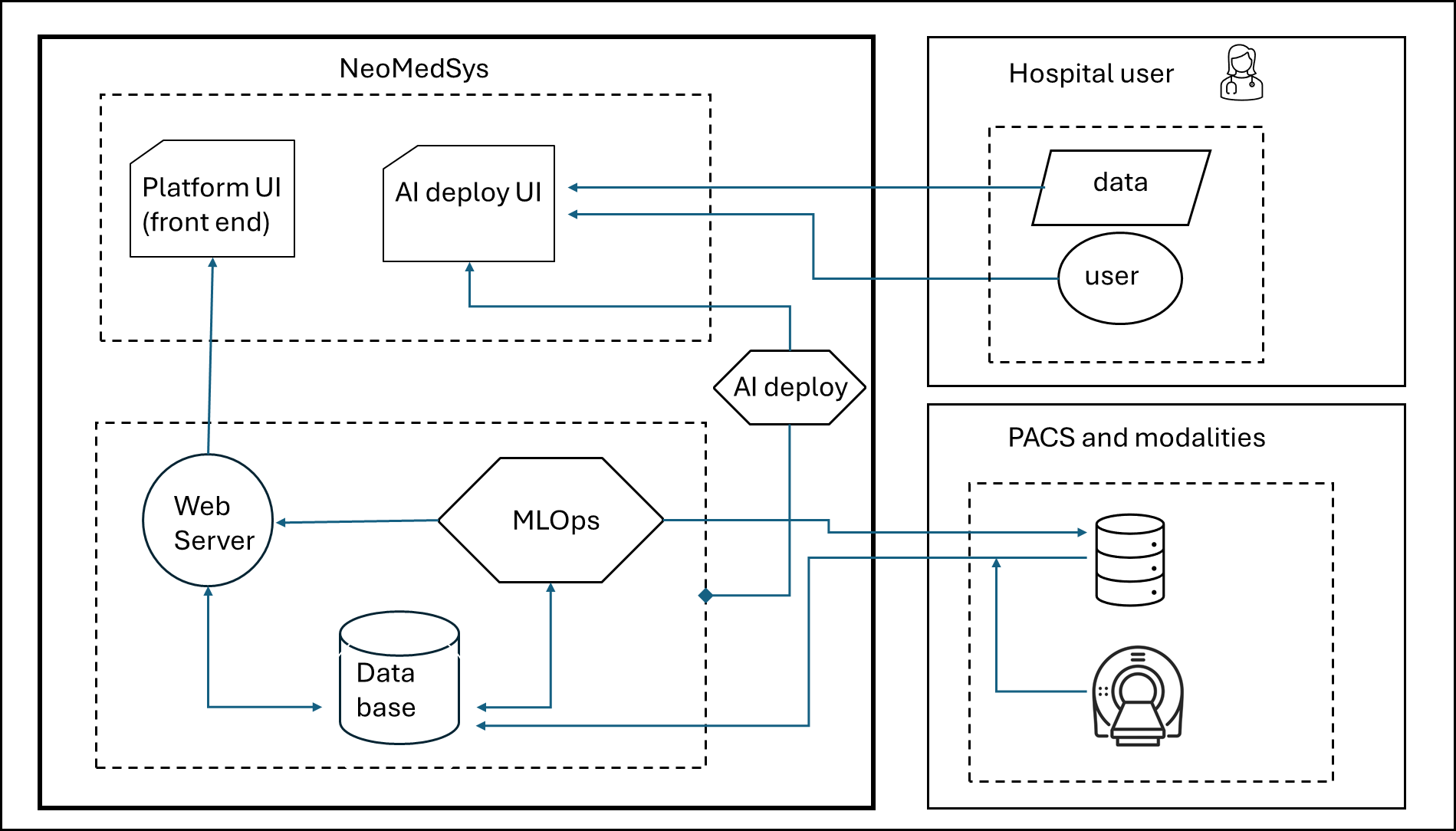}}
    \caption{The NeoMedSys on-premise system design overview. Medical imaging data is transmitted securely from PACS or scanning stations using the DICOM SEND protocol and ingested into the system. Data is processed and made available for AI workflows, model training, and inference. Authorized users can securely upload and deploy AI models within the platform. The deployment system integrates models into NeoMedSys's secure runtime environment. Inference pipelines and deployed/published AI models process data and can transmit AI-generated results back to PACS over the DICOM protocol.
}%
    \label{fig:neomed_arch}
\end{figure}

NeoMedSys is designed as a cloud-native hierarchical microservice, addressing both adaptability and security. The system design comprises three main system components: 
\begin{itemize}
    \item Machine Learning Operations (MLOps): This component is designed to manage the complete lifecycle of developing an AI prototype, with data security and training efficiency as priorities. Rigorous MLOps implementation is essential to ensure the safe development and deployment of AI models in a clinical environment. On the NeoMedSys platform, MLOps supports the following lifecycle stages:
    \begin{itemize}
        \item Data curation: Facilitating availability, organization, labeling, and correction of data to ensure it is well-structured and actionable for training and analysis.
        \item Model Development and fine-tuning: Enabling the creation and optimization of AI models tailored to any specific clinical needs.
        \item Deployment and integration: Allowing full-scale implementation with hospital radiology system (PACS).
        \item Monitoring: Providing tools to ensure continuous performance and compliance in real-world settings.
    \end{itemize}
    \item The database management framework. This component uses a robust cloud-native architecture that delivers enterprise-level data integrity, redundancy, failover, and scalability, designed specifically for the demands of clinical data management. The database framework is composed of distinct technology components, to ensure efficient handling of diverse medical data types, including structured and unstructured data and relationships between entities. 
    \item The NeoMedSys frontend provides a secure and intuitive environment for all user interactions. Built with a modern web architecture, it delivers high performance and responsiveness, continuously refined through clinical and research user feedback (Figure \ref{fig:neomed_arch}).
\end{itemize}

\begin{figure}[!htp]
    \centerline{\includegraphics[width=\textwidth ]{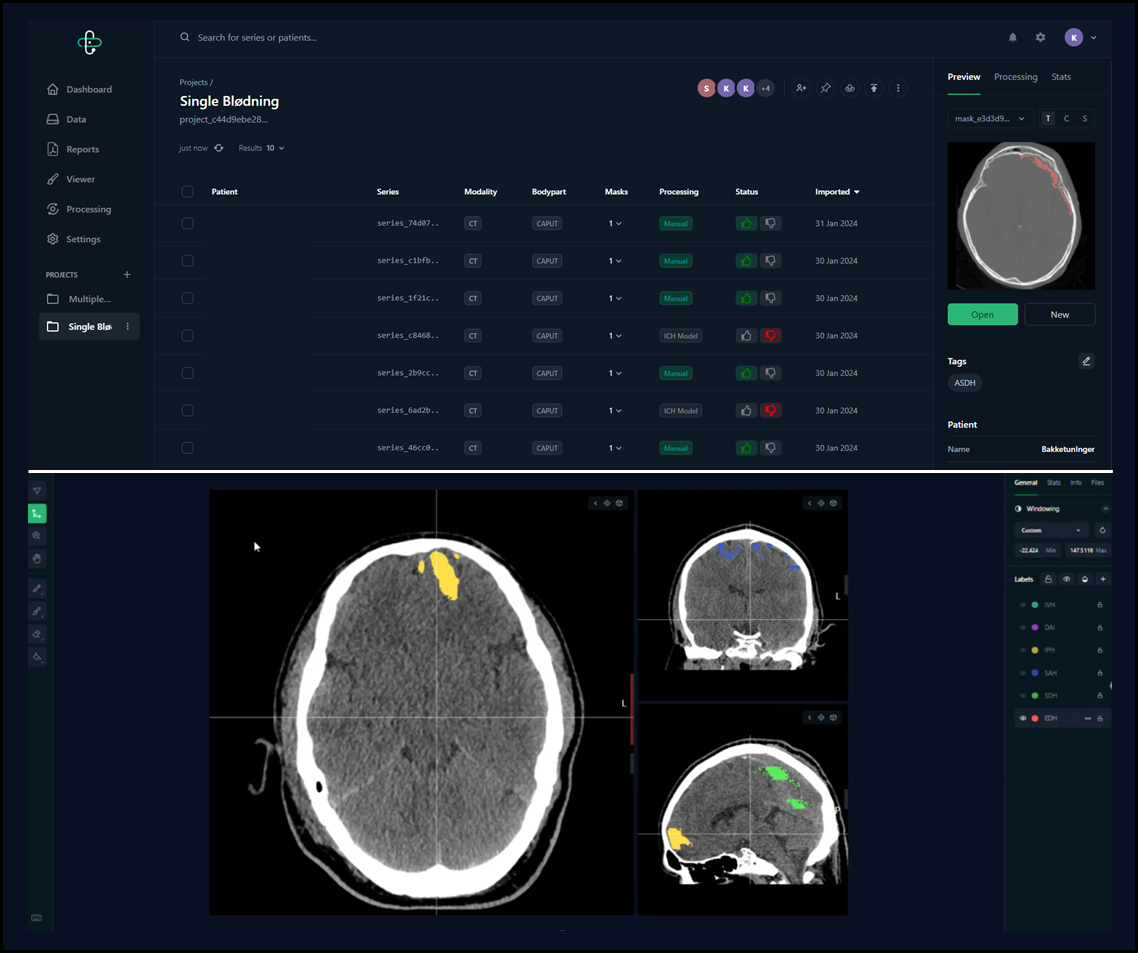}}
    \caption{The NeoMedSys platform user interface (UI)  (top) and NeoSeg UI for image viewing, annotation and segmentation touch-ups.}%
    \label{fig:neomed_ui}
\end{figure}

\begin{figure}[!htp]
    \centerline{\includegraphics[width=\textwidth ]{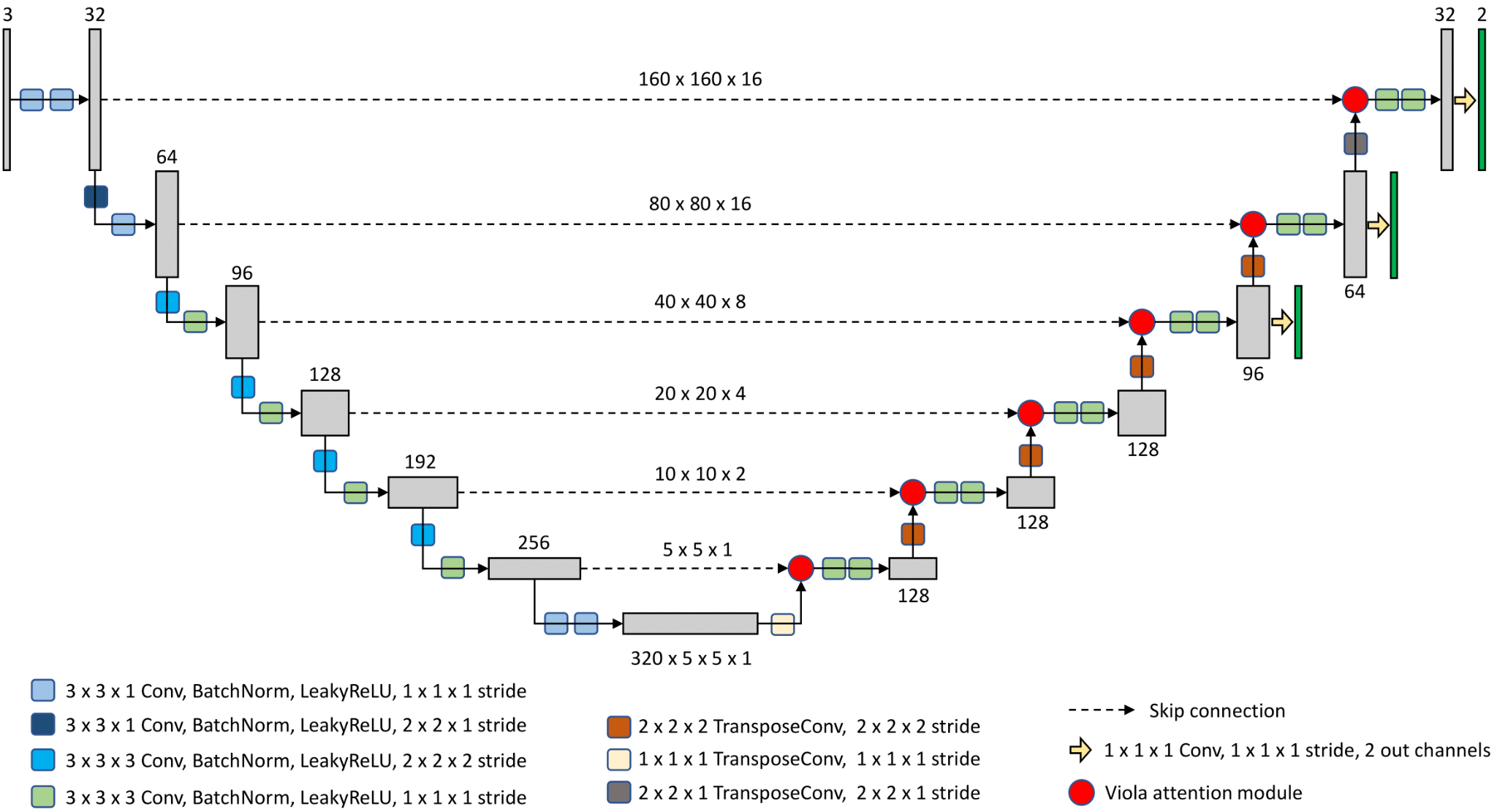}}
    \caption{VIOLA-AI model architecture powered by the proposed Voxels Intersecting along Orthogonal Levels Attention (VIOLA) module.}%
    \label{fig:viola_arch}
\end{figure}

\subsection{Data Security and Privacy}
To ensure patient privacy and data protection, all medical images transmitted from PACS to the NeoMedSys platform via the DICOM SEND protocol are de-identified at the source. The system’s cloud-native architecture adopts a zero-trust security model with rigorous network and access policy controls. Data integrity is maintained through a robust database management framework, and all information—including image metadata and AI outputs—is safeguarded by server-side encryption. This approach is designed to meet or exceed current best practices for healthcare data security and to ensure confidentiality and compliance with relevant privacy regulations.

\subsection{Real-world AI Model Refinement Process}

\begin{figure}[!htp]
    \centerline{\includegraphics[width=\textwidth ]{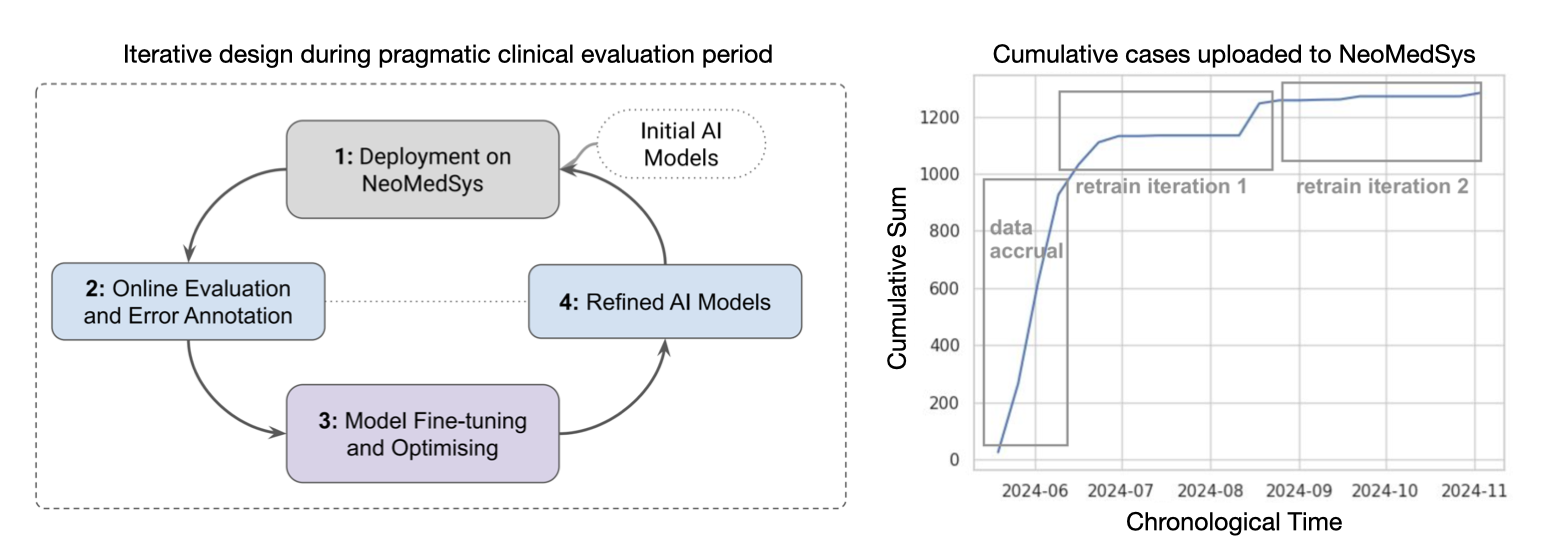}}
    \caption{(a) Continuous AI-Model Optimization Framework, (b) Timeline and cumulative cases for data ingestion, and iterative retraining.}%
    \label{fig:refine_process}
\end{figure}

The NeoMedSys platform facilitates the continuous optimization of the VIOLA-AI model for intracranial hemorrhage (ICH) detection through a real-world, data-driven approach. The VIOLA-AI model is an ensemble of two types of 3D neural networks, i.e., nnU-Net \cite{isensee_nnu-net_2021} and Voxel Intersecting along Orthogonal-Levels Attention U-Net (ViolaU-Net) \cite{liu_voxels_2023} architectures (Figure \ref{fig:viola_arch}).  This iterative refinement process, illustrated in Figure \ref{fig:refine_process}, ensures that the model's accuracy and adaptability are consistently enhanced within the dynamic clinical environment.
\begin{itemize}
    \item \textbf{Deployment on NeoMedSys:} Both the initially trained and refined versions of the VIOLA-AI model are deployed on the NeoMedSys platform to analyze incoming CT scans from patients with head trauma or stroke. Predictions are visualized through NeoMedSys’s integrated medical image viewer, providing radiologists with immediate access to the AI-generated outputs for clinical assessment.
    \item \textbf{Online Evaluation and Error Annotation:} Radiologists actively monitor VIOLA-AI's predictions in real-time through NeoMedSys (Figure \ref{fig:neomed_ui}), identifying and annotating specific errors such as false positives, false negatives, and segmentation inaccuracies.  They provide detailed annotations directly on the NeoMedSys platform, refining or generating segmentation masks as needed. These refined annotations are then incorporated into subsequent training iterations, enhancing the model’s detection and segmentation accuracy. 
    \item \textbf{Model Fine-tuning and Optimising:} Cumulative In-house data (sourced externally from NeoMedSys)  and the refined annotations dataset are systematically incorporated into retraining and fine-tuning datasets. This data-driven approach enables the model to learn from real-world examples and adapt to the specific characteristics of the local patient population. To further enhance accuracy, robustness, and the balance between sensitivity and specificity, techniques such as ensemble methods (averaging, voting, weighted ensembles), Test-Time Augmentation (TTA), and threshold calibration are employed.
    \item \textbf{Refined VIOLA-AI model:} The top-performing models, selected based on key performance metrics such as Dice, AUC, sensitivity, and specificity—evaluated on the local hold-out test set—are seamlessly deployed or updated on NeoMedSys for clinical integration. This streamlined deployment process ensures that the latest advancements are rapidly incorporated into the workflow, enhancing diagnostic precision and supporting radiologists with continuously improving AI-driven insights. Note that the model’s predictions are based on a default probability threshold of 0.5, a commonly used value for balancing sensitivity and specificity in binary classification. The NeoMedSys allows radiologists to manually adjust this threshold, enabling clinical customization.
\end{itemize}

\subsection{Datasets}
The training and test of VIOLA-AI leveraged a combination of in-house and public datasets, specifically INSTANCE \cite{li_state---art_2023} and BHSD \cite{wu_bhsd_2023}. Below is a detailed breakdown of the datasets used:
\begin{itemize}
    \item \textbf{Public Datasets:}
    \begin{itemize}
        \item INSTANCE: Contributed 100 bleed-positive cases for initial model training.
        \item BHSD: Provided an additional 152 bleed-positive cases for model training and 35 cases specifically allocated for the hold-out test set.
    \end{itemize}
    \item \textbf{In-House Datasets:}
    \begin{itemize}
        \item Initial In-House Data: collected from local hospitals (but not from NeoMedSys), comprising 241 bleed-positive cases with an average age of 55.1 years (±25.4 years), ranging from 3 to 94 years. 206 cases were used for training, while 35 cases were allocated for the hold-out test.
        \item New Annotations Data: Consisted of 40 newly refined annotated bleed-positive cases by radiologists on the NeoMedSys platform during online evaluation. Average age: 62.3 years (±21.2 years), ranging from 37 to 96 years.
        \item Data annotations were performed by HR, a 6-year experienced radiologist, under the supervision of KS, a senior neuroradiologist with more than 10 years of experience, both from Oslo University Hospital. Additional error feedback during the study period was also provided by HR and KS. While inter-rater reliability was not formally assessed, the consistent involvement of the same experts ensured consistency in the annotation and feedback process.
    
    \end{itemize}
    \item \textbf{Test Datasets:}
    \begin{itemize}
        \item Hold-out Test Set: This set comprised 70 bleed-positive cases (35 from in-house data and 35 from the public BHSD dataset) and 150 bleed-negative cases (from the in-house Negative Test Set). The average age was 51.8 years (±22.4 years), ranging from 8 to 96 years. This set was strictly reserved for final model evaluation and was not used during training or iterative refinement. It served as a benchmark to measure the model's performance and for model selection in the model deployment process.
        \item Online Test Set: This set included 154 bleed-positive cases and 150 bleed-negative cases (from the in-house Negative Test Set), used for the prospective, real-time evaluation of VIOLA-AI performance within the NeoMedSys platform during the study period. Average age: 60.9 years (±23.3 years), ranging from 6 to 96 years. This set was crucial for observing the model's behavior in a realistic clinical workflow.
        \item Negative Test Set: As noted earlier, this set comprised 150 bleed-negative cases sourced from in-house data, specifically selected as hard cases—those previously prone to false positives in earlier models during online testing.  The average patient age was 59.1 years (±24.3 years), ranging from 18 to 96 years. These cases were shared across both the hold-out and online test sets to ensure a consistent assessment of the model's specificity across both test environments.
    \end{itemize}
\end{itemize}
\begin{figure}[!htp]
    \centerline{\includegraphics[width=0.9\textwidth ]{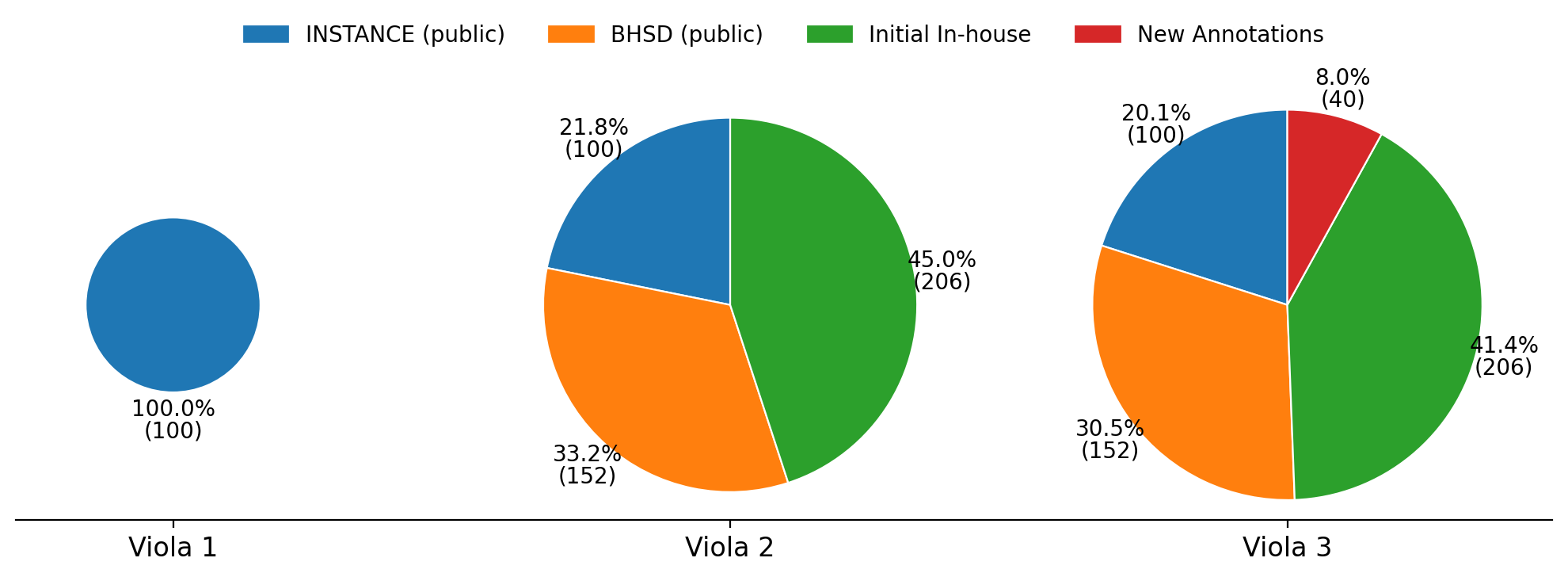}}
    \caption{Dataset used for training each VIOLA-AI Version: Viola-1 was trained on 100 cases from the public INSTANCE dataset \cite{li_state---art_2023}. Viola-2 was generated by incorporating an additional 152 cases from the public BHSD\cite{wu_bhsd_2023} dataset together with 206 initial in-house training cases. Viola-3 was further refined with 40 newly annotated local cases. These constitute the two main retraining rounds.}%
    \label{fig:data}
\end{figure}
VIOLA-AI underwent iterative improvements through increased training data as illustrated in Figure \ref{fig:data} Version 1 (Viola 1) was primarily trained on the INSTANCE public dataset. Version 2 (Viola 2) incorporated the BHSD public dataset and the initial in-house dataset. Version 3 (Viola 3) further incorporated the refined new annotations data in addition to previous datasets.

\section{Results}\label{result}

During the three-month study period, 2,167 patients underwent acute head CT scans, from 6 different CT scanners at two locations, to rule out ICH. The patient age range was 6 to 96 years, and 36.5\% were female. A total of 154 cases of ICH were confirmed by radiology reports. Other pathologies included non-hemorrhagic stroke, cerebral tumors, and infections. The mean time from CT examination to the radiology report was 35 minutes (range: 3 minutes to 5 hours), while the mean time for VIOLA-AI result generation within NeoMedSys was 3 minutes.

The performance of VIOLA-AI versions 1, 2, and 3 were evaluated on a hold-out dataset of 220 cases (70 positive and  150 negative) and an online evaluation set of 304 cases (154 positive and 150 negative) respectively. The results are summarized in Table \ref{tab1} and Table \ref{tab2}.

\begin{table}[h]
\caption{Summarizing local Hold-out test performance}\label{tab1}%
\begin{tabular}{@{}cccccccc@{}}
\toprule
VIOLA-AI & Dice  & Sens & Spec & AUC  & Accu & Preci & F1\\
\midrule
Viola 1 & 0.449   & 0.842   & 0.847  & 0.903  & 0.818   & 0.670  & 0.802 \\
Viola 2 & 0.656   & 0.986  & \textbf{0.940}  & 0.984  & 0.918   & 0.802  & 0.911 \\
Viola 3 & \textbf{0.666}    & \textbf{1.0}   & 0.90  & \textbf{0.994}  & \textbf{0.927}   & \textbf{0.814}  & \textbf{0.921} \\
\botrule
\end{tabular}
\end{table}
\begin{table}[h]
\caption{Summarizing Online Test Performance}\label{tab2}%
\centering
\begin{tabular}{@{}ccccccc@{}}
\toprule
VIOLA-AI & Sens & Spec & AUC  & Accu & Preci & F1\\
\midrule
Viola 1  & 0.792  & 0.807  & 0.873  & 0.799   & 0.808  & 0.799 \\
Viola 2   & 0.831  & 0.887  & 0.913  & 0.855   & 0.877  & 0.855 \\
Viola 3    & \makecell{\textbf{0.903} \\ (+7.2\%)}  
           & \makecell{\textbf{0.893} \\ (+0.6\%)}  
           & \makecell{\textbf{0.949} \\ (+3.6\%)}  
           & \makecell{\textbf{0.898} \\ (+4.3\%)}  
           & \makecell{\textbf{0.897} \\ (+2.0\%)}  
           & \makecell{\textbf{0.898} \\ (+4.3\%)} \\
\botrule
\end{tabular}
\end{table}

Significant performance improvements were observed in VIOLA-AI between versions 1 and 3, attributable to radiologist feedback. This was evident in both the hold-out test and online test, demonstrating the effectiveness of the continuous learning and iterative refinement methodology. A particularly notable enhancement occurred in Viola 3, where the addition of 40 new annotation cases during fine-tuning of Viola 2 resulted in an increase in AUC from 0.913 to 0.949 (+3.6\%) and an increase in sensitivity from 0.831 to 0.903 (+7.2\%).
\begin{figure}[!htp]
    \subfloat[]{%
      \includegraphics[clip,width=\textwidth]{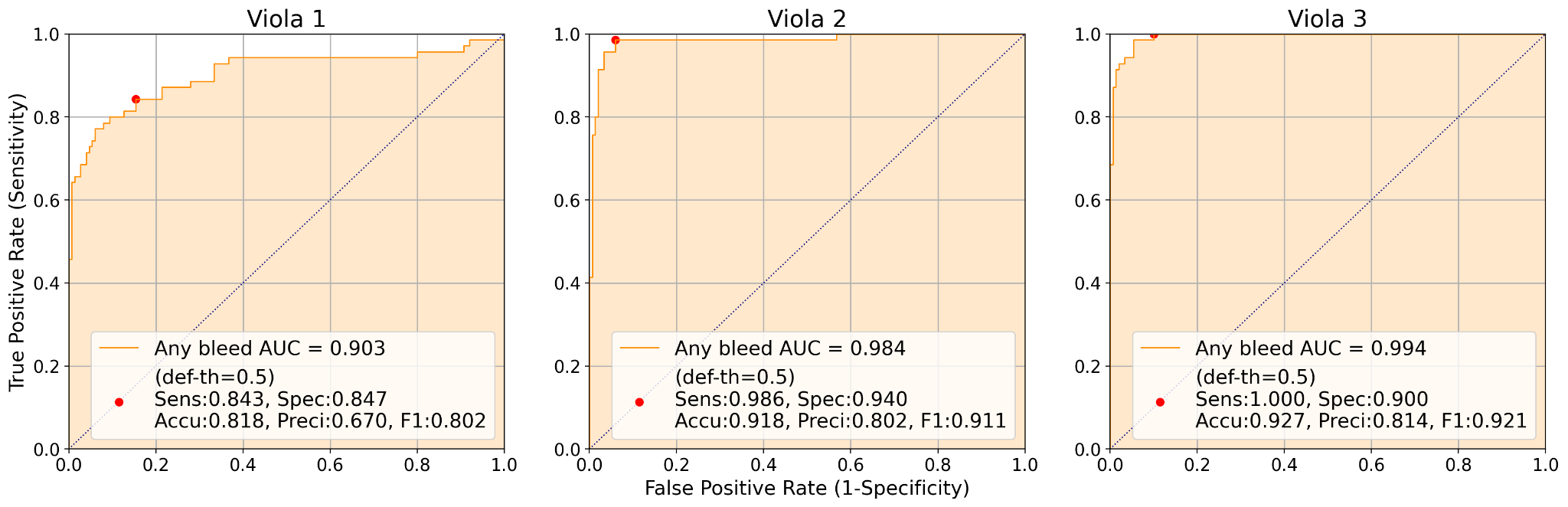}%
    }
    
    \subfloat[]{%
      \includegraphics[clip,width=\textwidth]{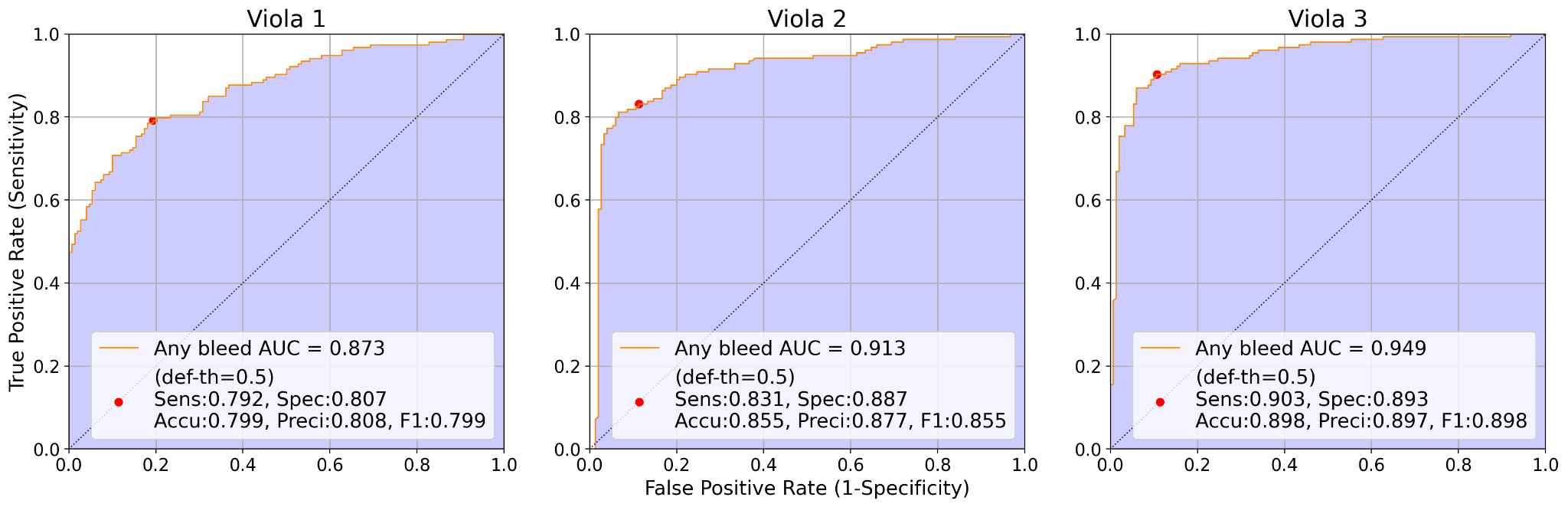}%
    }
    \caption{ROC curves on the hold-out test (a) and Online test (b) respectively.}%
    \label{fig:roc}
\end{figure}

\begin{figure}[!htp]
    \centerline{\includegraphics[width=0.8\textwidth ]{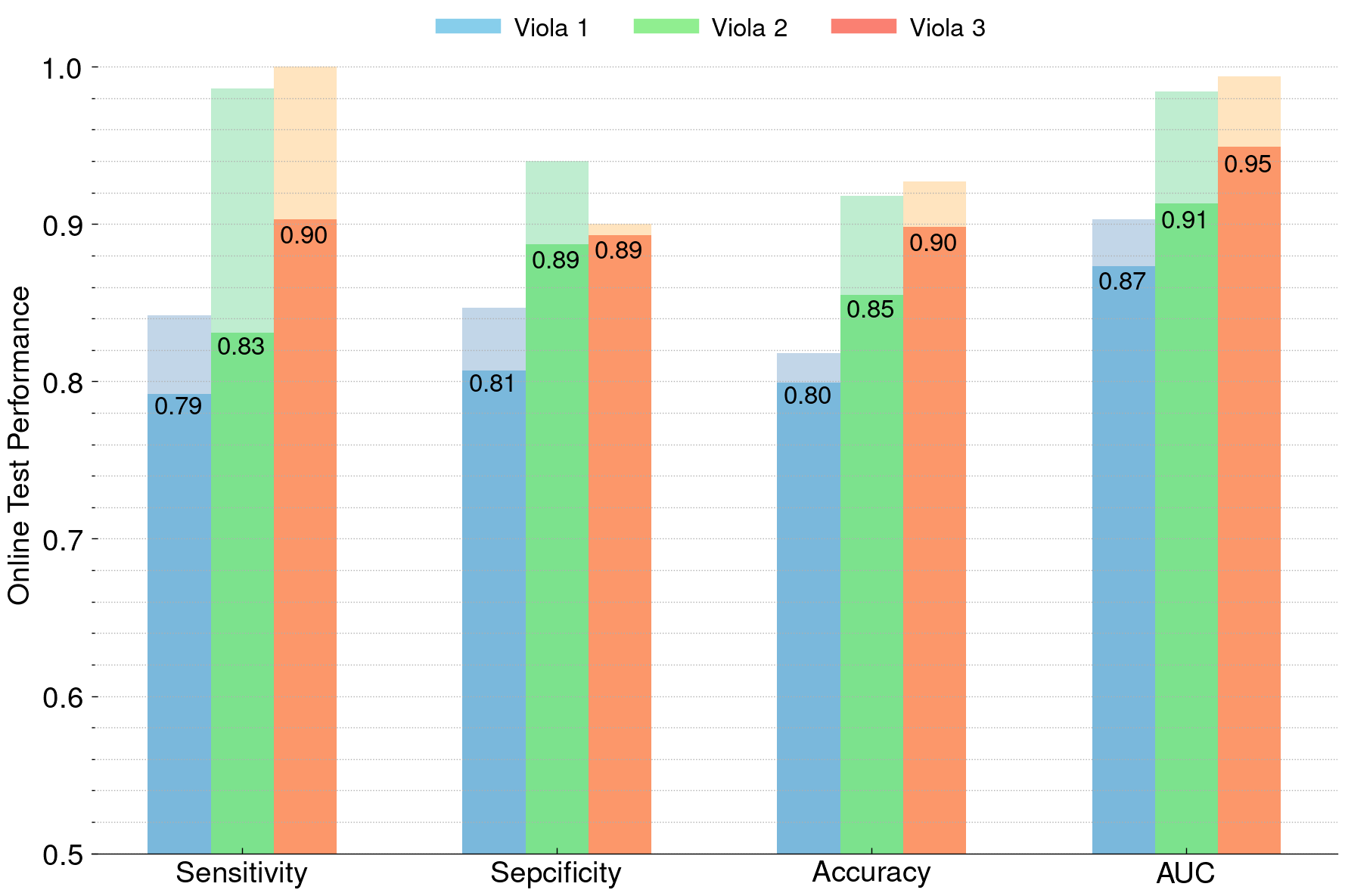}}
    \caption{A bar chart comparing key evaluation metrics—Sensitivity, Specificity, Accuracy, and AUC—across three VIOLA-AI models (Viola 1, Viola 2, and Viola 3) in an online test setting. Lighter-colored segments represent performance gaps compared to the local hold-out test for reference.}%
    \label{fig:bar_compa}
\end{figure}

Figure \ref{fig:roc} illustrates the ROC curves for the hold-out test (a) and online test (b), and Figure \ref{fig:bar_compa} visually compares online evaluation results on three key metrics (sensitivity, specificity, and AUC) across the three VIOLA-AI versions, further demonstrating the enhanced performance via valuable error feedback iteration refinement.

\begin{figure}[!htp]
    \centerline{\includegraphics[width=\textwidth ]{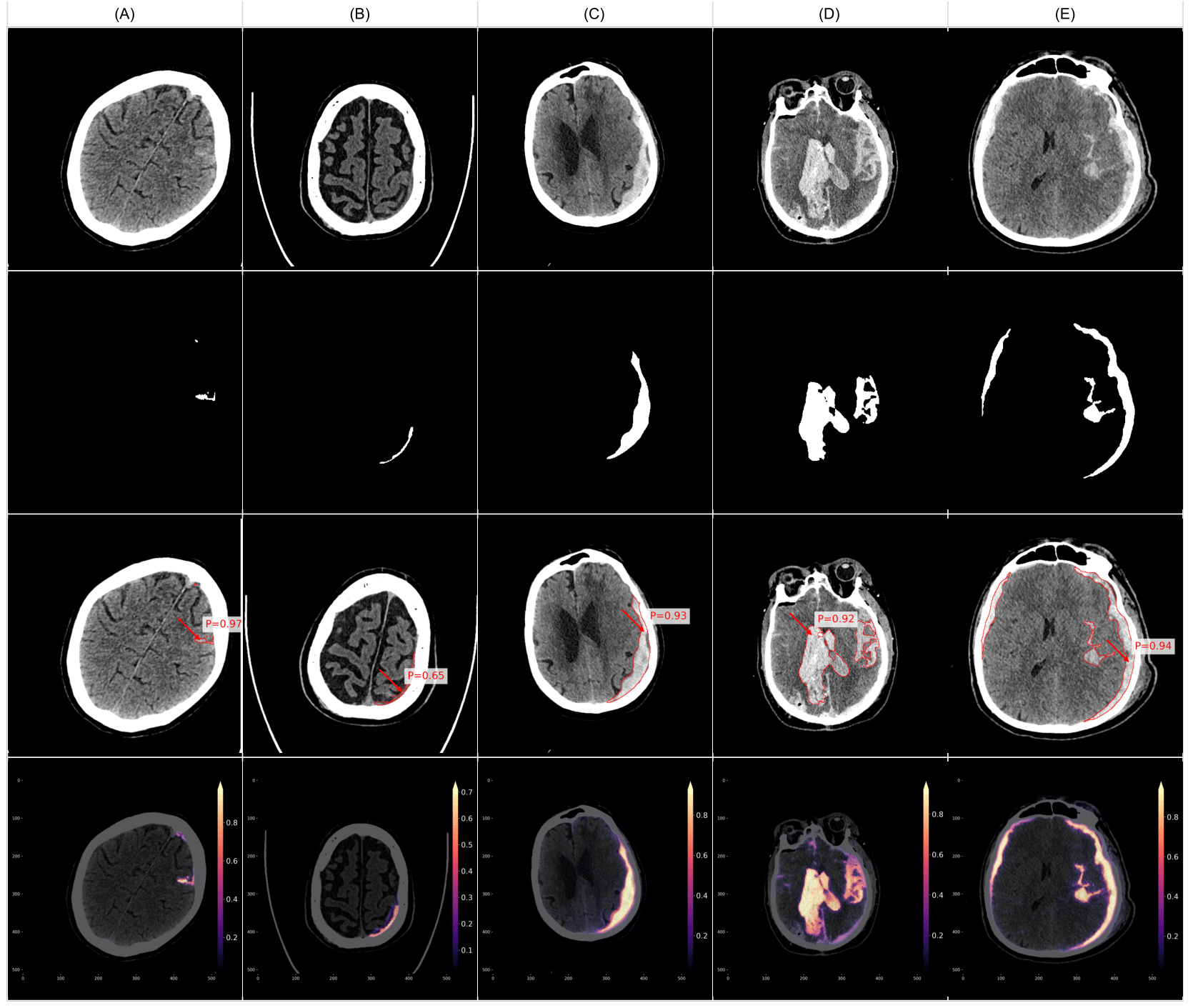}}
    \caption{True positive cases. First row: raw axial CT slices of the brain showing various cases with suspected small hemorrhage. Second row: Segmentation masks providing clear delineation of the hemorrhage extent. Third row: Overlaying the detected bleeding regions on the raw slices with red outlines indicating the segmented areas and calibrated confidence scores (P-values) are annotated which reflects the model’s probability estimation for bleed presence. Fourth row: providing a probability heatmap visualization where color intensity represents the likelihood of bleed presence in different regions. Warmer colors (yellow/red) indicate higher confidence while cooler colors (purple/blue) indicate lower probabilities.}%
    \label{fig:true_pos}
\end{figure}

\begin{figure}[!htp]
    \centerline{\includegraphics[width=\textwidth ]{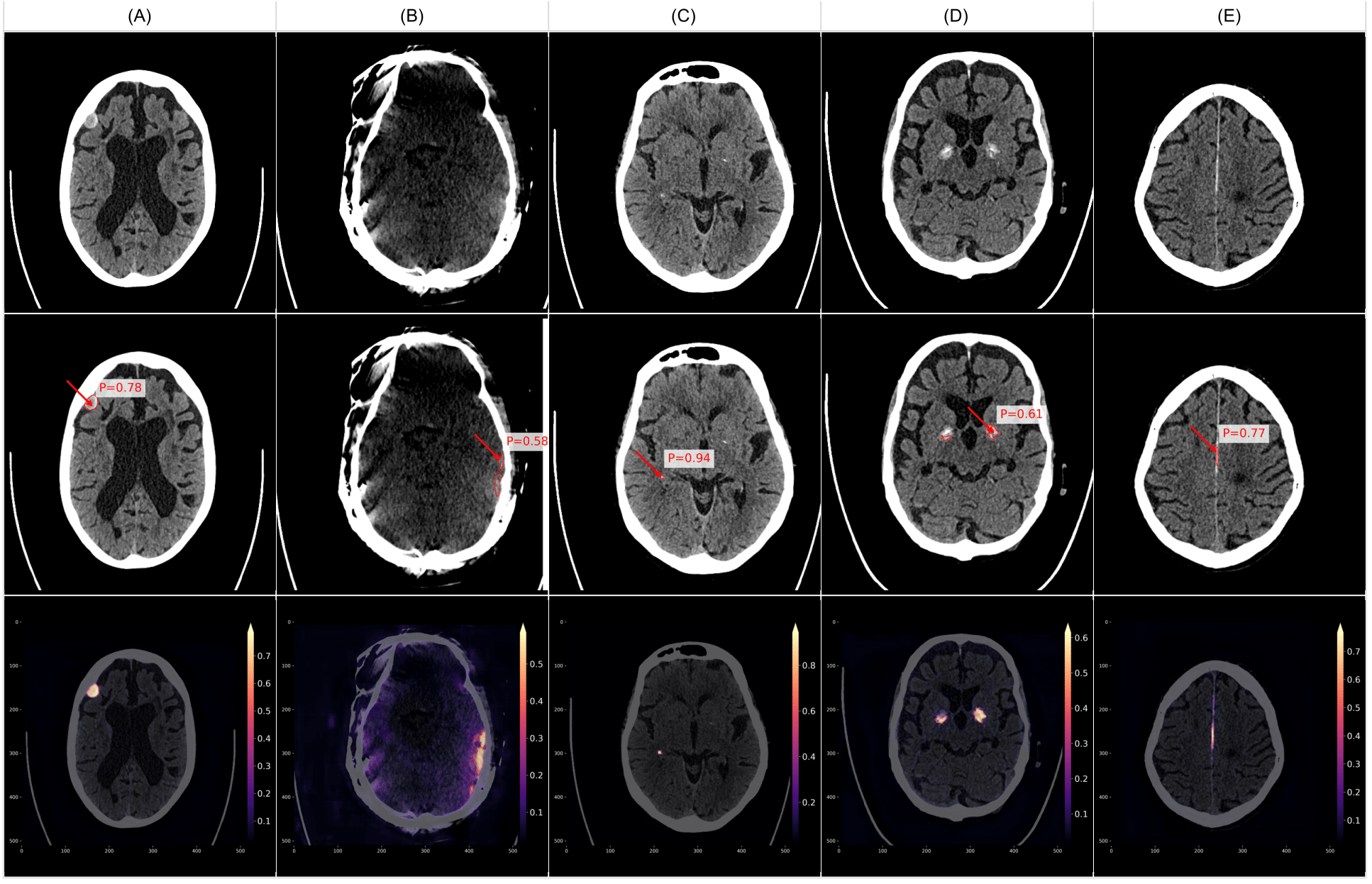}}
    \caption{False positive cases. Top row: raw axial CT slices. Middle row: detected hemorrhage regions (red outlines) on the raw slices with prob-values annotated. Bottom row: associated probability heatmaps. (A) Partly calcified meningioma, (B) motion artifacts, (C) calcification in association with choroid plexus, (D) basal ganglia calcifications, and (E) falx and dura calcifications.}%
    \label{fig:false_pos}
\end{figure}

\begin{figure}[!htp]
    \centerline{\includegraphics[width=\textwidth ]{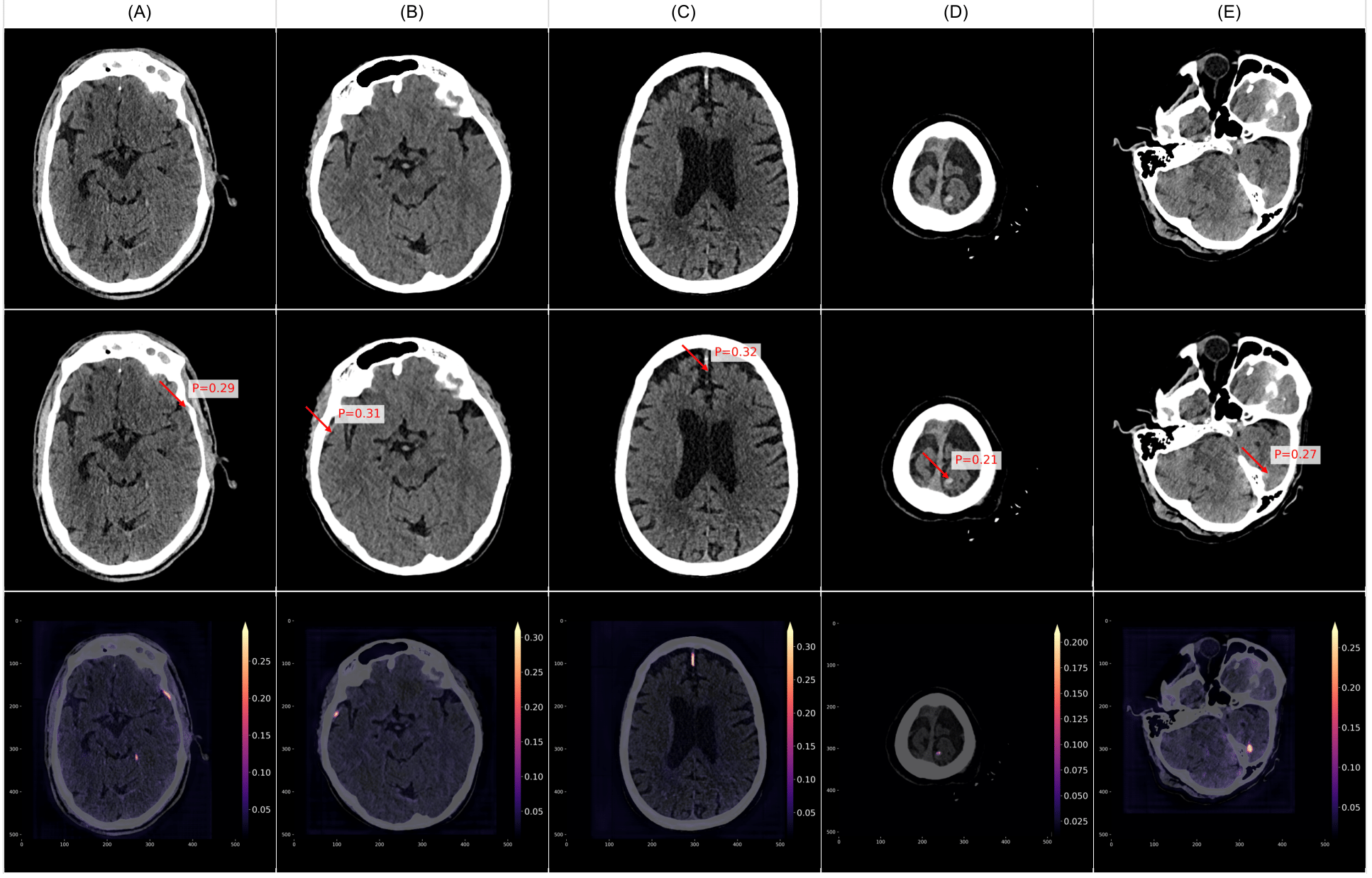}}
    \caption{False negative cases. Top row: raw axial CT slices. Middle row: detected hemorrhage regions (red outlines) on the raw slices with prob-values annotated. Bottom row: associated probability heatmaps. Note that our model did identify regions of potential hemorrhage, but the predicted probability values (e.g. from 0.2-0.4) were below the default detection threshold of 0.5, leading to their exclusion from positive classifications.}%
    \label{fig:false_neg}
\end{figure}

To provide a more intuitive understanding of VIOLA-AI's performance, we have selected several representative cases for illustration, i.e., True positive case examples (Figure \ref{fig:true_pos}), False positive case examples (Figure \ref{fig:false_pos}), and False negative case examples (Figure \ref{fig:false_neg}). 

Figure \ref{fig:true_pos} presents examples of true positive cases where VIOLA-AI successfully detected intracranial hemorrhage in CT scans. Specifically, (A) and (B) illustrate smaller ICH, (C) depicts acute SDH, (D) shows a combination of SAH, IPH, and IVH, and (E) highlights both acute SDH and SAH. These images showcase the model's capability to accurately detect a range of hemorrhage types and volumes, underscoring its high sensitivity and precision in identifying positive cases.

Figure \ref{fig:false_pos} illustrates examples of false positive cases, where VIOLA-AI incorrectly identified ICH in CT scans that did not contain any bleeds. These misclassifications often arise because some bleed-like features closely resemble non-hemorrhagic structures. Contributing factors include imaging artifacts (e.g., B: motion artifacts) and the model’s heightened sensitivity to certain high-density regions, such as various calcifications. Examples include A: a partly calcified meningioma, C: choroid plexus calcifications, D: basal ganglia calcifications, and E: falx and dura calcifications. Analyzing these false positives is crucial for refining the model to better distinguish between true hemorrhages and similar-looking non-bleed findings, thereby improving overall accuracy.

Figure \ref{fig:false_neg} presents examples of false negative cases, where VIOLA-AI failed to detect intracranial hemorrhage (ICH) in CT scans despite the presence of actual bleeds. Although VIOLA identified regions with potential hemorrhage, the predicted bleed probability values fell below the default detection threshold of 0.5, resulting in their exclusion from positive classifications. These false negatives often involve subtle or small bleeds, lesions in anatomically challenging regions, or bleed types that the model struggles to recognize. Addressing these cases is crucial for enhancing the model’s sensitivity, reducing missed diagnoses, and improving patient safety.

\section{Discussion}\label{discc}
In this study, we have shown a tangible example where a deep learning ICH segmentation tool was evaluated in a clinical setting and NeoMedSys enabled an end-to-end platform to ingest new cases directly from PACS, expert segmentation touch-ups, and model retraining. We found that sensitivity, specificity, and AUC reached an acceptable level of performance after two rounds of VIOLA-AI retraining and over three months of a pragmatic clinical evaluation. Compared to other AI models detecting and annotating ICH on head CT, the results from VIOLA-AI are slightly short of the pooled sensitivity and specificity of 92\% and 94\% from a recent meta-analysis  \cite{maghami_diagnostic_2023}. The increased performance of the 3rd version where it used 40 cases of local segmentations including false negatives from previous versions demonstrated in Figure \ref{fig:data}. The latter highlights the importance of the integrative relationship between VIOLA-AI, NeoMedSys and clinical systems. Very few platforms provide the opportunity to use local data to both refine data with redefining segmentations in an intuitive manner to further facilitate retraining and upgrading the algorithm. Using in-house datasets to train models provides a valuable feature that in-house models can easily benefit from compared to commercial models. Some commercial models are poorly documented via published peer-reviewed papers allowing them to provide undocumented polished results \cite{van_leeuwen_artificial_2021}. As demonstrated in Figure \ref{fig:bar_compa} when results are tested on local hold-out dataset the results improve considerably whilst tested on a more clinically equivalent dataset the results are more realistic. As reported by papers introducing commercial models the accuracy features drop when introduced into a clinical workflow \cite{omoumi_buy_2021}.

Well known challenging factors for both radiologists and AI models are locating and diagnosing; small ICH like SAH and thin SDH right adjacent to the skull shown as examples in Figure \ref{fig:true_pos}, but again with error feedback the in-house models can be adjusted to and upgraded especially with an interactive platform like NeoMedSys. To have a high sensitivity often produces a number of false positives, some examples presented in Figure \ref{fig:false_pos}, in this dataset the most common presentations were either other pathologies such as tumors or calcifications, or small locations such as close to venous sinuses or calcified/thick dura in the midline. The two latter also pose challenges to radiologists, hence such false positives could be of benefit to bring focus to these areas. False negatives (Figure \ref{fig:false_neg}) are always worrisome when it comes to ICH. Detecting bleeds that need acute intervention such as surgery is not difficult for either AI or radiologists to detect, however, the challenge lies with the smaller bleeds, especially the small SAH where studies have shown that they pose no immediate threat to patients, diagnosing these at a later stage does not pose a risk \cite{rosen_routine_2018}. As false negatives exist eliminates the fact of AI models reliably producing final reports. While our primary focus was overall ICH detection, we acknowledge that a more granular analysis across specific ICH subtypes (e.g., SAH, SDH vs. EDH) will be important in future work as our dataset expands. Furthermore, our deliberate use of a dedicated 'Negative Test Set' enriched with hard cases was a key strategy to address data imbalance and improve specificity, as reflected in the performance gains from VIOLA-1 to VIOLA-3.

Whenever implementing AI models and platforms several quality management systems and other regulatory aspects need to be addressed and accounted for \cite{kim_holistic_2024}. Even though tedious to implement, commercial products with relevant approvals are deemed easier to incorporate into a clinical workflow as a decision tool. In-house developed models due to regulatory processes can be challenging to implement as a decision AI tool, however as shown in this study in-house when used as an assisting tool when having a resource-sufficient AI department with direct communication with the radiology department can be even more fruitful as this set-up promotes continuous refinements and upgrades.  

Others have proposed similar academic implementations. Hawkings et al suggested a system for in-clinic AI deployment using a customized open-source XNAT instance combined with NVIDIA’s Clara Deployment framework \cite{hawkins_implementation_2023}. However, their framework did not address model retraining and refinements via feedback from end-users. The NeoMedSys platform supports responsible AI adoption by enhancing interpretability, trust, and regulatory alignment. Probability heatmaps and segmentation masks help clinicians understand model outputs, while the human-in-the-loop design builds trust by allowing radiologists’ feedback to directly improve performance across versions. By positioning the AI as an assisting tool for research and quality management—rather than a final decision-maker—the platform provides a practical pathway for integration within current regulatory frameworks.

The importance of the medical field taking a larger role in AI development has been shown to be fruitful to increase use clinically \cite{najjar_redefining_2023, wiggins_imaging_2021, geis_ethics_2019, dantonoli_ethical_2020, kundisch_deep_2021, warman_using_2024, seyam_utilization_2022-1, allen_jr_road_2019, dikici_integrating_2020}. An extremely important role for radiologists and doctors to grab as larger bodies express radiologists as a dying breed even as commercial providers sell their AI applications as adjunct and assisting to radiologist-led interpretation rather than replacing radiologists \cite{van_leeuwen_how_2022}. With the latter hybrid approach, studies show improved diagnostic accuracy \cite{van_leeuwen_how_2022}. Larger institutions with the infrastructure and clinical and technological competence should invest in a multidisciplinary involvement for the development and implementation of both locally developed algorithms and platforms to bridge the gap from research to the clinical setting. While a direct, head-to-head comparison with other commercial or publicly available SOTA models was beyond the scope of this pragmatic evaluation, the main strength of the NeoMedSys platform lies in enabling rapid, iterative model refinement adapted to local patient populations and clinical workflows. For context, Qure.ai's qER \footnote{\url{https://www.qure.ai/}}, FDA-approved in 2020, has reported sensitivities of 81\%–93.6\% and specificities of 89.7\%–94\% \cite{mabit2025real}. Similarly, Aidoc \footnote{\url{https://www.aidoc.com/}}, another FDA-cleared decision-support tool for ICH detection, has been validated with sensitivities of 88.7\%–96.2\% and specificities of 92.3\%–99.0\% \cite{seyam2022utilization}. The final version of our VIOLA-AI model achieved a sensitivity of 90.3\% and a specificity of 89.3\% overall, demonstrating that an iterative, in-house refinement process can reach performance levels comparable to those of established commercial systems.

It is important to discuss the current findings and NeoMedSys platform in the context of some limitations. First, although information material was provided and there was buy-in by the staff members working in the diagnostic imaging sites, there were communication gaps. This resulted in many head CT scans being performed but not pushed to NeoMedSys, however this was pushed at a later date so all images were analyzed VIOLA-AI. The decision to include a human-in-the-loop to push cases to NeoMedSys was a deliberate choice as it ensured that the local sample would be an appropriate choice for the local cohort, thereby avoiding incorrect imaging series or patient populations. Second, the 3-month pragmatic study duration was chosen as the reasonable duration to capture many cases but not overwhelm the staff that volunteered to push cases to NeoMedSys. It is intriguing to consider alternative trial designs, which would be the topic of future work, such as longer in duration or including an element of randomization (i.e. week on / week off). Finally, another limitation of this study is its single-country setting, which may increase the risk of overfitting to local data characteristics. Future work will extend the NeoMedSys platform to support multi-site deployments and external validation, thereby enabling a more rigorous assessment of generalizability across diverse clinical settings and patient populations. The platform’s capacity to capture imaging protocol metadata, such as scanner type and calibration versions, will further facilitate analysis of performance variability.

\section{Conslusions}

The NeoMedSys platform has proven to be an effective tool for bridging the gap between AI research and clinical practice in radiology. By enabling real-time validation and iterative refinement of AI models, the platform facilitates the integration of AI into routine clinical workflows. The successful deployment and refinement of VIOLA-AI for ICH detection demonstrate the platform's potential to improve diagnostic accuracy and efficiency.

The results of this pragmatic clinical evaluation show that iterative retraining, driven by radiologist feedback, led to substantial improvements in VIOLA-AI's performance. The final version of the model achieved high sensitivity, specificity, and AUC, indicating its clinical utility. These findings emphasize the importance of continuous learning and adaptation for AI models in real-world settings.

While this study focused on a single AI model and pathology, the NeoMedSys platform is designed to be scalable and adaptable to various imaging modalities and clinical conditions.  Future developments will expand NeoMedSys to support additional AI applications with multi-site deployments and external validation, incorporate automated retraining features, and enhance scalability to larger clinical settings, ultimately contributing to safer and more efficient patient care. By continuing to refine and expand the NeoMedSys platform, we aim to contribute to safer, more efficient, and more accurate patient care.

\bmhead{Abbreviations}
\begin{abbreviations}
\item[CT] Computed Tomography
\item[VIOLA] Voxel Intersecting along Orthogonal-Levels Attention
\item[AI] Artificial Intelligence
\item[AUC] Area Under the receiver operating characteristic Curve 
\item[ROC] Receiver Operating Characteristic
\item[ICH] 	Intracranial Hemorrhage
\item[SAH]	Subarachnoid Hemorrhage
\item[SDH]	Subdural Hematoma
\item[EDH]	Epidural Hematoma
\item[IPH]	Intraparenchymal Hemorrhage
\item[IVH]	Intraventricular Hemorrhage
\item[MLOps] Machine Learning Operations
\item[UI]	User interface
\end{abbreviations}

\bmhead{Acknowledgement}
The authors wish to thank the radiographers at the different sites, Oslo Legevakt and Ullevål Hospital with a key function of “pushing” head CTs to the NeoMedSys platform. We thank Espen Kristoffersen and Per Selnes at Akershus University Hospital for providing annotated data used in the refinement training of the VIOLA-AI model.

\bmhead{Funding}
The authors acknowledge support from the Helse Sør-Øst regional health authority of Norway (Grant Number: 2021057 and 2017073), and the Research Council of Norway (Grant Number: 325971).

\bmhead{Data Availability Statement}
Aspects of the data for this study will be made available upon request. The VIOLA-AI model and trained weights are freely available for use as a docker image and can be accessed from the following GitHub account: \url{https://github.com/samleoqh/Viola-Unet}. The tool runs in a docker environment and the trained model weights can be downloaded. An open version of the NeoMedSys platform to be used with anonymized data can be accessed at \url{https://www.neomedsys.io}.

\bmhead{Conflict of Interest}
The authors declare that the research was conducted in the absence of any commercial or financial relationships that could be construed as a potential conflict of interest.

\bmhead{Competing interests}
The authors declare no competing interests.

\bmhead{Author Contributions}
QL developed the AI tool, contributed data analysis and drafted the manuscript. JN developed the NeoMedSys platform and helped with study design and writing. HR contributed to the study design, data annotation, data collection and interpretation. EM and MR helped with AI model deployment on the NeoMedSys and data collection. BJM helped with the study design, and drafted the manuscript. AB contributed the study design, interpretation and writing.  KS contributed to the study design, data collection, interpretation, and writing. All authors contributed to the article and approved the submitted version.






\bibliography{sn-bibliography}

\end{document}